\pdfoutput=1

\documentclass[11pt]{article}

\usepackage[]{acl}

\usepackage{times}
\usepackage{latexsym}

\usepackage[T1]{fontenc}

\usepackage[utf8]{inputenc}

\usepackage{microtype}

\usepackage{inconsolata}

\usepackage{bm}      
\usepackage{mathtools}
\usepackage{amsmath}
\usepackage{amsfonts} 

\usepackage[ruled, lined, linesnumbered, commentsnumbered, longend]{algorithm2e}
\usepackage{xcolor}

\SetCommentSty{mycommfont}

\DeclareMathOperator*{\argmax}{arg\,max}

\newcommand{\figref}[1]{Figure~\ref{fig:#1}}
\newcommand{\tblref}[1]{Table~\ref{table:#1}}

\newcommand{\eqnref}[1]{Equation~\ref{eq:#1}}

\newcommand{\B}[1]{\textbf{#1}}
\newcommand{\U}[1]{\underline{#1}}
\newcommand{\I}[1]{\textit{#1}}

\title{Low-resource neural machine translation with morphological modeling}

\author{Antoine Nzeyimana \\
  University of Massachusetts Amherst \\
  \texttt{anthonzeyi@gmail.com} \\}

\begin{document}

\maketitle

\begin{abstract}

Morphological modeling in neural machine translation (NMT) is a promising approach to achieving open-vocabulary machine translation for morphologically-rich languages. However, existing methods such as sub-word tokenization and character-based models are limited to the surface forms of the words. In this work, we propose a framework-solution for modeling complex morphology in low-resource settings. A two-tier transformer architecture is chosen to encode morphological information at the inputs. At the target-side output, a multi-task multi-label training scheme coupled with a beam search-based decoder are found to improve machine translation performance. An attention augmentation scheme to the transformer model is proposed in a generic form to allow integration of pre-trained language models and also facilitate modeling of word order relationships between the source and target languages. Several data augmentation techniques are evaluated and shown to increase translation performance in low-resource settings. We evaluate our proposed solution on Kinyarwanda $\leftrightarrow$ English translation using public-domain parallel text. Our final models achieve competitive performance in relation to large multi-lingual models. We hope that our results will motivate more use of explicit morphological information and the proposed model and data augmentations in low-resource NMT.

\end{abstract}

\section{Introduction}

Neural Machine Translation (NMT) has become a predominant approach in developing machine translation systems. Two important innovations in recent state-of-the-art NMT systems are the use of the Transformer architecture~\cite{vaswani2017attention} and sub-word tokenization methods such as byte-pair encoding (BPE)~\cite{sennrich2015neural}. However, for morphologically-rich languages(MRLs), BPE-based tokenization is only limited to the surface forms of the words and less grounded on exact lexical units (i.e. morphemes), especially in the presence morphographemic alternations~\cite{bundy1984morphographemics} and non-concatenative morphology~\cite{kastner2019non}. In this work, we tackle the challenge of modeling complex morphology in low-resource NMT and evaluate on Kinyarwanda, a low-resource and morphologically-rich language spoken by more than 15 million people in Eastern and Central Africa\footnote{\url{https://en.wikipedia.org/wiki/Kinyarwanda}}.

To model the complex morphology of MRLs in machine translation, one has to consider both source-side modeling (i.e. morphological encoding) and target-side generation of inflected forms (i.e. morphological prediction). We explicitly use the morphological structure of the words and the associated morphemes, which form the basic lexical units. For source-side encoding, morphemes are first produced by a morphological analyzer before being passed to the source encoder through an embedding mechanism. On the target side, the morphological structure must be predicted along with morphemes, which are then consumed by an inflected form synthesizer to produce surface forms. Therefore, this approach enables open vocabulary machine translation since morphemes can be meaningfully combined to form new inflected forms not seen during training. 

Previous research has shown that certain adaptations to NMT models, such as the integration of pre-trained language models~\cite{zhu2020incorporating, sun2021multilingual}, can improve machine translation performance. We explore this idea to improve low-resource machine translation between Kinyarwanda and English. Our model augmentation focuses on biasing the attention computation in the transformer model. Beside augmentation from pre-trained language model integration, we also devise an augmentation based solely on the word order relationship between source and target languages. These model augmentations bring substantial improvement in translation performance when parallel text is scarce.

One of the main challenges facing machine translation for low-resource languages obviously is parallel data scarcity. When the training data has limited lexical coverage, the NMT model may tend to hallucinate~\cite{raunak2021curious,xu2023understanding}. Additionally, for a morphology-aware translation model, there is a problem of misaligned vocabularies between source and target languages. This makes it harder for the model to learn to copy unknown words and other tokens that need to be copied without translation such as proper names. 
To address these challenges, we take a data-centric approach by developing tools to extract more parallel data from public-domain documents and websites. 
We also use various data augmentation techniques to increase lexical coverage and improve token copying ability where necessary. By combining these data-centric approaches with our morphology-aware NMT model, we achieve competitive translation performance in relation to larger multi-lingual NMT models. To have a comprehensive evaluation, we evaluate our models on three different benchmarks covering different domains, namely Wikipedia, News and Covid-19.

In short, our contribution in this work can be summarized as follow:
\begin{itemize}
    \item We propose and evaluate methods for source-side and target-side morphological modeling in neural machine translation of morphologically-rich languages.
    \item We propose a generic method for attention augmentation in the transformer architecture, including a new cross-positional encoding technique to fit word order relationships between source and target language.
    \item We evaluate on Kinyarwanda$\leftrightarrow$English translation across three benchmarks and achieve competitive performance in relation to existing large multi-lingual NMT models.
    \item We release tools for parallel corpus construction from public-domain sources and make our source code publicly available to allow reproducibility\footnote{\url{https://github.com/anzeyimana/KinMT_NAACL2024}}.
\end{itemize}

\section{Methods}

Machine translation (MT) can be considered as the task of accurately mapping a sequence of tokens (e.g. phrase, sentence, paragraph) in the source language $S = (s_1, s_2, ... s_n)$ to a sequence of tokens in the target language $T = (t_1, t_2, ... t_m)$ with the same meaning. The learning problem is then to estimate a conditional probability model that produces the optimal translation $T^*$, that is: 

\vspace{-0.14in}
\begin{equation*}
    T^* = \argmax_T P(T|S,T_{<};\Theta),
\vspace{-0.15in}
\end{equation*}

where $T_{<}$ accounts for the previous output context and $\Theta$ are parameters of the model (that is a neural network in the case of NMT).

In this section, we describe our model architecture as an extension of the basic Transformer architecture~\cite{vaswani2017attention} to enable morphological modeling and attention augmentation. We also describe our data-centric approaches to dataset development and augmentation in the context of the low-resource Kinyarwanda$\leftrightarrow$English machine translation.

\subsection{Model architecture}
\label{sec:arch}

\begin{figure}[!ht]
 \centering
    \includegraphics[scale=0.42]{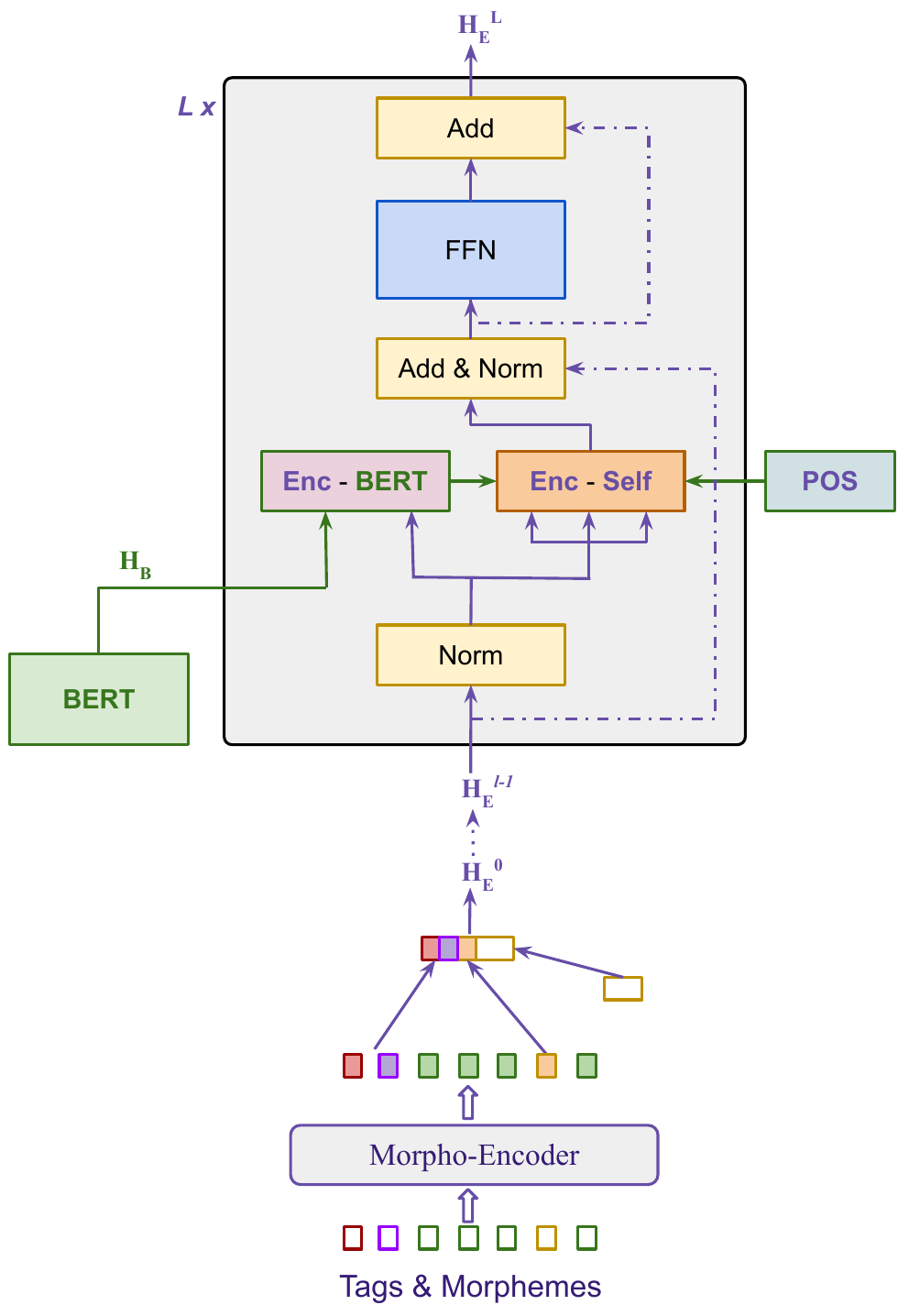}
 \captionof{figure}{\label{fig:encoder_architecture} Encoder architecture}
 \label{fig:encoder}
\end{figure}

\begin{figure}[!ht]
 \centering
    \includegraphics[scale=0.42]{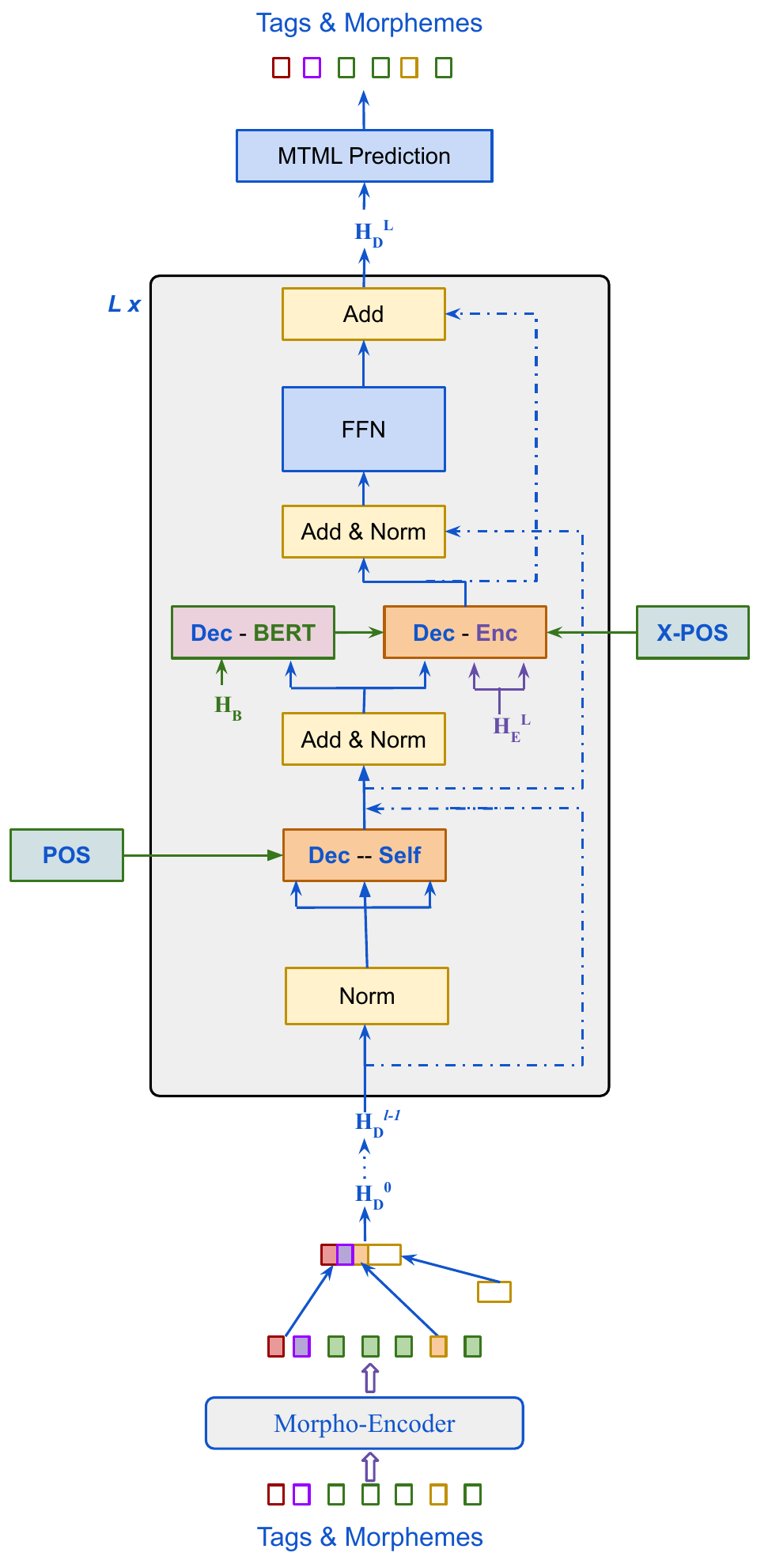}
 \captionof{figure}{\label{fig:decoder_architecture} Decoder architecture}
 \label{fig:decoder}
\vspace{-0.2in}
\end{figure}

The transformer architecture~\cite{vaswani2017attention} for machine translation uses a multi-layer bidirectional encoder to process source language input, and then feeds to an auto-regressive decoder to produce the target language output. Our adaptation of the transformer encoder is depicted in \figref{encoder} while the decoder is shown in \figref{decoder}. They both use pre-LayerNorm configuration~\cite{nguyen2019transformers} of the transformer.

The \B{attention module} of the transformer architecture is designed as querying a dictionary made of key-value pairs using a softmax function and then projecting a weighted sum of value vectors to an output vector, that is:
\vspace{-0.1in}
\begin{equation}
    Attention(Q,K,V) = Softmax(\frac{QK^T}{\sqrt{d}})V,
\label{eq:1}
\end{equation}
where $K$,$Q$,$V$ are projections of the hidden representations of inputs at a given layer. Given a hidden representation of a token $v_i$ attending to a sequence of tokens with hidden representations $(w_1, w_2, ... w_n)$, the output of the attention module corresponding to $v_i$ can be formulated as:
\begin{equation}
\begin{split}
v^{'}_i=\sum_{j=1}^n \frac{\exp (\alpha_{ij})}{\sum_{j'=1}^n\exp (\alpha_{ij'})}(w_jW_{V}), \\
\text{where the logits } \alpha_{ij}=\frac{1}{\sqrt{d}} (v_iW_{Q})(w_jW_{K})^T,
\end{split}
\label{eq:2}
\end{equation}
with $W_Q\in \mathbb{R}^{d \times d_K}, W_K\in \mathbb{R}^{d \times d_K},$ and $W_V\in \mathbb{R}^{d\times d_V}$ being learnable projection matrices. $d, d_K$ and $d_V$ are the dimensions of the input, key and value respectively.

\subsubsection{Attention augmentations}
\citet{ke2020rethinking} proposed to add bias terms to the logits $\alpha_{ij}$ in \eqnref{2} as untied positional encoding, disentangling a mixing of token and position embeddings. We generalize this structure by allowing more biases to be added to the logits $\alpha_{ij}$ in \eqnref{2}.

Specifically, we explore augmenting two attention components in the transformer architecture by making the following extensions:
\begin{enumerate}
    \item \I{For \B{Source-to-source self-attention} at each encoder layer}: We integrate embeddings from a pre-trained BERT~\cite{devlin2019bert} model. This adds rich contextual information as BERT models are pre-trained on large monolingual data and perform well on language understanding tasks. We also add positional encodings at this level, similar to~\cite{ke2020rethinking}. Therefore, the logits $\alpha_{ij}$ at encoder layer $l$ become:
    \item[]
    \begin{equation}
      \begin{aligned}
        \alpha_{ij} = \frac{1}{\sqrt{3d}} [ & (x_i^{(l)}W_Q^{(l)})(x_j^{(l)}W_K^{(l)})^T \\
        \quad   & +(p_iU_Q)(p_jU_K)^T + r_{j-i}\\
        \quad   & +(x_i^{(l)}V_Q^{(l)})(b_jV_K^{(l)})^T],
      \end{aligned}
    \end{equation}
    where $x_i^{(l)}$ and $x_j^{(l)}$ are hidden representations of source tokens at positions $i$ and $j$ respectively of the encoder layer $l$; $p_i$ and $p_j$ are absolute position embeddings; $r_{j-i}$ is a relative position embedding; $b_j$ is a pre-trained BERT embedding of token at position $j$ and $W_Q^{(l)}, W_K^{(l)}, U_Q, U_K, V_Q^{(l)}$ and $V_K^{(l)}$ are learnable projection matrices. We note that this formulation requires the source encoder to match the same token vocabulary as the BERT embedding model.

    \item \I{For \B{target-to-source cross-attention} at each decoder layer $l$}, we also augment the attention logits with pre-trained BERT embeddings of the source sequence. Additionally, we propose a new type of embedding: \B{cross-positional embeddings}. These are embeddings that align target sequence positions to input sequence positions. Their role can be thought as of learning word order relationships between source and target languages. Their formulation is closely similar to the untied positional encoding proposed by~\cite{ke2020rethinking}, but they cross from target to source positions, thus, we name them cross-positional (XPOS) encodings. The attention logits $\alpha_{ij}^{'}$ at this level thus become:
    \begin{equation}
      \begin{aligned}
        \alpha_{ij}^{'} = \frac{1}{\sqrt{3d}} [ & (y_i^{(l)}W_Q^{'(l)})(x_j^{(L)}W_K^{'(l)})^T \\
        \quad   & +(p_i^{'}U_Q^{'})(p_j^{'}U_K^{'})^T  + r_{j-i}^{'}\\
        \quad   & +(y_i^{(l)}V_Q^{'(l)})(b_jV_K^{'(l)})^T],
      \end{aligned}
    \end{equation}
    where $y_i^{(l)}$ is the hidden representation of the target token at position $i$ and $x_j^{(L)}$ is the hidden representation of source token at position $j$ of the final encoder layer $L$. $p_i^{'}$ and $p_j^{'}$ are absolute target and source XPOS embeddings, $r_{j-i}^{'}$ is a target-to-source relative XPOS embedding, $b_j$ is a pre-trained BERT embedding of source token at position $j$ and $W_Q^{'(l)}, W_K^{'(l)}, U_Q^{'}, U_K^{'}, V_Q^{'(l)}$ and $V_K^{'(l)}$ are the learnable projection matrices.
\end{enumerate}

\subsubsection{Morphological encoding}
For most transformer-based encoder-decoder models, the first layer inputs are usually formed by mapping each sub-word token, such as those produced by BPE, to a learnable embedding vector. However, BPE-produced tokens do not always carry explicit lexical meaning. In fact, they cannot model non-concatenative morphology and other morpho-graphemic processes as these BPE tokens are solely based on the surface forms. Inspired by the work of~\citet{nzeyimana2022kinyabert}, we explore using a small transformer encoder to form a word-compositional model based on the morphological structure and the associated morphemes.

Depicted at the input layers in~\figref{encoder} and~\figref{decoder}, the morphological encoder or \textit{Morpho-Encoder} is a small transformer encoder that processes a set of four embedding units at the word composition level: (1) \B{the stem}, (2) a \B{variable number of affixes}, (3) a coarse-grained \B{part-of-speech (POS) tag} and, (4) a fine-grained \B{affix set index}. An affix set represents one of many frequent affix combinations observed empirically. Thus, the affix set index is equivalent to a fine-grained \B{morphological tag}.

The \textit{Morpho-Encoder} processes all word-compositional units as a set without any ordering information. This is because none of these units can be repeated in the same word. In cases of stem reduplication phenomena~\cite{inkelas2000reduplication}, only one stem is used, while the reduplication structure is captured by the affix set. At the output of the \textit{Morpho-Encoder}, hidden representations corresponding to units other than the affixes are pulled and concatenated together to form a word hidden vector to feed to the main sequence model. In addition to this, a new stem embedding vector at the sequence level is also concatenated with the pulled vectors from the \textit{Morpho-Encoder} to form the final hidden vector representing the word.

In our experiments, we use 24,000 most frequent affix combinations as affix sets. Any infrequent combination of affixes can always be reduced to a frequent one by removing one or more affixes. However, all affixes still contribute to the word composition via the \textit{Morpho-Encoder}.

We note that the \textit{Morpho-Encoder} applies to both the encoder and the auto-regressive decoder's input layers for all types of tokens. While the word-compositional model above relates mostly to inflected forms, we are able generalize this to other typed of tokens such as proper names, numbers and punctuation marks. For these other tokens, we process them using BPE and consider the resulting sub-word tokens as special stems without affixes.

\subsubsection{Target-side morphology learning}
\label{sec:target_learn}
Considering the morphological model employed at the input layer, the decoder outputs for a morphologically-rich target language must be used to predict the same types of morphological units used at the input layer, namely, the stem, affixes, the POS tag and the affix set. This becomes a multi-task and multi-label (MTML) classification problem which requires optimizing multiple objectives, corresponding to 4 types loss functions:
\begin{equation}
    \begin{aligned}
    \ell_{S} & = \ell_{CE}(f_{S}(h^L), y_{S}) \\
    \ell_{A}^{(i)} & = \ell_{BCE}(f_{A}(h^L), y_{A}^{(i)}),  \forall i \in \mathcal{A} \\
    \ell_{P} & = \ell_{CE}(f_{P}(h^L), y_{P}) \\
    \ell_{AS} & = \ell_{CE}(f_{AS}(h^L), y_{AS}),
    \end{aligned}
\end{equation}
where $h^L$ is the decoder output, $\mathcal{A}$ is the set of affix indices, $f_S, f_{A}, f_P, $ and $f_{AS}$ are prediction heads transforming the decoder output to probabilities over the sets of stems, affixes, POS tags and affix sets  respectively. $y_S, y_A, y_P, $ and $y_{AS}$ respectively correspond to the stem, affixes, POS tags and affix set of a target word $y$. $\ell_{CE}$ is a cross-entropy loss function and $\ell_{BCE}$ is a binary cross-entropy loss function.

A naive approach to the MTML problem consists of summing up all the losses and optimizing the sum. However, this can lead to a biased outcome since individual losses take on different ranges and have varying levels of optimization difficulty. Complicating the problem further is the fact that individual objectives can contribute conflicting gradients, making it harder to train the multi-task model with standard gradient descent algorithms. A potential solution to this problem comes form the multi-lingual NMT literature with a scheme called Gradient Vaccine~\cite{wang2020gradient}. This method attempts to mediate conflicting gradient updates from individual losses by encouraging more geometrically aligned parameter updates. We evaluate both the naive summation and the Gradient Vaccine methods in our experiments.

\subsubsection{Morphological inference}
The decoder architecture presented in subsection \ref{sec:target_learn} only predicts separate probabilities for stem, POS tag, affix set and affixes. But the translation task must produce surface forms to generate the output text. The challenge of this task is that greedily picking the items with maximum probability may not produce the best output and may even produce incompatible stems and affixes, that is, we must produce a stem and affixes of the same inflection group (e.g. verb, noun, pronoun, etc..). It is also known that beam search algorithm generally produces better sequence outputs than greedy decoding. Therefore, we design an adaptation of the beam search algorithm, where at each step, we produce a list of scored candidate surface forms together with their morphological information to feed back to the decoder's input. The design criteria is to make sure the top predicted items can form compatible pairs of stems and affix sets. The main requirement for the algorithm is the availability of a morphological synthesizer that can produce surface forms given an inflection group, the stem and compatible affixes. The morphological synthesizer must also respect all existing morpho-graphemic rules for the language. We we provide detailed pseudocode for the decoding algorithm in Appendix \ref{sec:morpho_decoding}. The algorithm has 4 basic steps: 
\begin{enumerate}
    \item voting on inflection group
    \item filtering out less probable stems and affixes
    \item selecting target affixes, and finally
    \item morphological synthesis for each final stem and affixes combination.
\end{enumerate}

\subsection{Dataset}

Dataset development and data-centric approaches to neural machine translation (NMT) are of paramount importance for low-resource languages. This is because the most limiting factor is the scarcity of parallel data. While describing the data collection process and pre-processing steps is important, it is equally important to fully disclose the data provenance as there are typically a limited number of sources of parallel data per language. We conduct our experiments using public-domain parallel text. In this section, we describe our parallel data gathering process as well as the reliable sources we used to source Kinyarwanda-English bitext. We also describe simple data augmentation techniques we used to boost the performance of our experimental models. Due to copyright and licensing restrictions, we cannot redistribute our experimental dataset. Instead, we release the tools used for their construction from original sources. The sizes of the parallel datasets we gathered are provided in Appendix~\ref{sec:data_summary}.

\subsection{Official Gazette}

Official gazettes are periodical government journals typically with policy and regulation content. When a country has multiple official languages, content may be available as parallel text with each paragraph of the journal available in each official language. We took this opportunity and collected an experimental parallel text from the Official Gazette of the Republic of Rwanda\footnote{\url{https://www.minijust.gov.rw/official-gazette}}, where Kinyarwanda, French, English and Swahili are all official languages. This is an important source of parallel text given that it covers multiple sectors and is usually written with high standards by professionals, part of a dedicated government agency.

The main content of Rwanda's official gazette is provided in a multi-column portable document format (PDF), mostly 3 columns for Kinyarwanda, English and French. In our experiments, we process page content streams by making low level modifications to Apache PDFBox Java library\footnote{\url{https://pdfbox.apache.org/}}, where the inputs are unordered set of raw characters with their X-Y page coordinates and font information. We track columns by detecting margins (by sorting X-coordinates of glyphs) and reconstruct text across consecutive pages. A key opportunity for parallel alignment comes from the fact that most official gazettes paragraphs are grouped by consecutive article numbers such as ``Ingingo ya 1/Article 1'', ``Ingingo ya 2/Article 2'', and so on. We use these article enumerations as anchors to finding parallel paragraphs across the three languages. A language identification component is also required to know which column correspond with which language as the column ordering has been changing over time.




\subsection{Jw.org website}

Jw.org website publishes religious and biblical teachings by Jehovah's Witnesses, with cross-references into multiple languages. While this website data has been used for low-resource machine translation before~\cite{agic2019jw300}, the isolated corpus is no longer available due to license restrictions. However, the content of the original website is still available to web browsers and crawlers. We take this advantage and gather data from the site to experiment with English$\leftrightarrow$Kinyarwanda  machine translation.


\subsection{Bilingual dictionaries}

Bilingual dictionaries are also useful for low-resource machine translation. While most of their parallel data are made of single words, they can still contribute to the translation task, albeit without any sentence-level context.
The 2006 version of the Iriza dictionary~\cite{habumuremyi2006} is generally publicly available in PDF format. Similar to the Official Gazette case, we use low level modifications to the Apache PDFBox library and extract dictionary entries grouped by a source word and a target synset. Another bilingual dictionary we used is \I{kinyarwanda.net} website\footnote{\url{https://kinyarwanda.digital/}}, which was developed by volunteers to help people learning Kinyarwanda or English. In addition to these bilingual dictionaries, we manually translated about 8,000 Kinyarwanda words whose stems could not be found in any of the existing parallel data sources. Some of these terms include recently incorporated but frequently used Kinyarwanda words such as loanwords and also alternate common spellings. Examples include words such as: `abazunguzayi' (\textit{hawkers}), `akabyiniro' (\textit{night club}), `canke' (from Kirundi: \textit{or}), `mitiweli' (from French: ``mutuelle santé''). Together with data from bilingual dictionaries, the above dataset forms a special training subset we call `lexical data', because it augments the lexical coverage of our main dataset. We evaluate its effectiveness in our experiments.




\subsection{Monolingual data}

Backtranslation~\cite{edunov2018understanding} is a proven technique for leveraging monolingual data in machine translation. We developed a corpus of Kinyarwanda text by crawling more than 200 websites and extracting text from several books to form a monolingual dataset to use for back-translation. The final corpus contains about 400 million words and tokens or 16 million sentences. We also formed an English text corpus of similar size by crawling eight major Rwandan and East African English newspapers (3.3 million sentences) in addition to Wikipedia English corpus (7.3 million sentences)\footnote{\url{https://www.kaggle.com/datasets/mikeortman/wikipedia-sentences}} and global English news data (5.4 million sentences)\footnote{\url{https://data.statmt.org/news-crawl/en/news.2020.en.shuffled.deduped.gz}}.


\subsection{Data augmentations}

Source to target copying in NMT is a desirable ability when faced with untranslatable terms such as proper names. However, when the source and target vocabularies are not shared, it is harder for the model to learn this ability.
In order to enforce this copying ability in our NMT model, we take a data-centric approach by including untranslatable terms in our dataset with the same source and target text. This augmentation includes the following datasets: (1) All numeric tokens and proper names from our Kinyarwanda monolingual corpus, (2) Names of locations from the World Cities dataset\footnote{\url{https://github.com/datasets/world-cities/blob/master/data/world-cities.csv}}, and (3) Names of people from CMU Names corpus~\cite{kantrowitz2018names} and the Names Dataset~\cite{NameDataset2021}.

We also add synthetic data for number spellings by using rule-based synthesizers to spell 200,000 random integers between zero and 999 billions. For Kinyarwanda side, we developed our own synthesizer, while we used \texttt{inflect} python package\footnote{\url{https://pypi.org/project/inflect/}} for English.

Code-switching is one characteristic of some low-resource languages such as Kinyarwanda. To cope with this issue, we add foreign language terms and their English translations to the training dataset for Kinyarwanda $\rightarrow$ English models as if the foreign terms were valid Kinyarwanda inputs. For this, we include all English phrases from \I{kinyarwanda.net} online dictionary, 100 popular French terms and 20 popular Swahili terms.




\section{Experiments}

\subsection{Experimental setup}

The model presented in section~\ref{sec:arch} was implemented from scratch using PyTorch framework~\cite{paszke2019pytorch} version 1.13.1. Our model hyper-parameters along with training and inference hyper-parameters are provided in appendix~\ref{sec:hyper_params}. Training was done using a hardware platform with 8 Nvidia RTX 4090 GPUs, with 256 gigabytes of system memory, on a Linux operating system. We used mixed precision training with lower precision in BFLOAT-16 format. For Kinyarwanda$\rightarrow$English model, one gradient update step takes 0.5 second and convergence is achieved after 40 epochs. For English$\rightarrow$Kinyarwanda model with Gradient Vaccine scheme, one gradient update step takes 1.1 seconds while convergence is achieved after 8 epochs.

In all our experiments on Kinyarwanda $\leftrightarrow$ English translation, only Kinyarwanda side (as source or target) is morphologically modelled, while the English side always uses sub-word tokenization. For Kinyarwanda source side with attention augmentation, we use a pre-trained BERT model similar to KinyaBERT~\cite{nzeyimana2022kinyabert} whose input units/token ids are the same as for the NMT encoder. In fact, this pre-trained KinyaBERT model has the same two-tier architecture as the NMT encoder. Therefore, they are aligned to the same words/tokens. We use a Kinyarwanda morphological analyzer\footnote{\url{https://github.com/anzeyimana/DeepKIN}} to perform both tokenization, morphological analysis and disambiguation~\cite{nzeyimana2020morphological}.

On English sides (either source or target), we do not perform morphological analysis and only use a standard single-tier transformer architecture. On the source side, we use a BPE-based tokenization and a corresponding pre-trained RoBERTA model provided by fairseq package~\cite{ott2019fairseq}. Similarly, on English target side, we use a BPE-based tokenization from a Transformer-based auto-regressive English language model~\cite{ng2019facebook} from the same fairseq package.


\begin{table*}[!ht]
\centering
\resizebox{0.98 \textwidth}{!}{
{\renewcommand{\arraystretch}{1.3}
\begin{tabular}{| c | c c | c c | c c | c c |}
\hline
~  & \multicolumn{2}{|c|}{FLORES-200} & \multicolumn{2}{|c|}{MAFAND-MT} & \multicolumn{2}{|c|}{TICO-19} & \multicolumn{2}{|c|}{\B{Average}} \\
\B{Use copy data?}  &  {BLEURT}  &  {ChrF++}  &  {BLEURT}  &  {ChrF++}  &  {BLEURT}  &  {ChrF++}  &  {BLEURT}  &  {CHR++} \\
\hline
No  &  55.5  &  39.2  &  52.3  &  37.1  &  52.2  &  33.4  &  53.3  &  36.6 \\
Yes  &  \B{56.8}  &  \B{40.3}  &  \B{54.8}  &  \B{39.6}  &  \B{52.9}  &  \B{34.0}  &  \B{54.9}  &  \B{38.0} \\
\hline
\end{tabular}
}
}
\caption{Impact of proper name copying ability: \B{Kinyarwanda $\rightarrow$ English}. Maximum scores are shown in bold.}
\label{table:name_copy}
\vspace{-0.05in}
\end{table*}


\begin{table*}[!ht]
\centering
\resizebox{0.98 \textwidth}{!}{
{\renewcommand{\arraystretch}{1.3}
\begin{tabular}{| l | c | c c | c c | c c | c c |}
\hline
{~}  &  \B{\#Params}  & \multicolumn{2}{|c|}{FLORES-200} & \multicolumn{2}{|c|}{MAFAND-MT} & \multicolumn{2}{|c|}{TICO-19} & \multicolumn{2}{|c|}{\B{Average}} \\
\multicolumn{1}{|c|}{\B{Setup}}  &  \B{ ($\times$ 1M)}  &  {BLEURT}  &  {ChrF++}  &  {BLEURT}  &  {ChrF++}  &  {BLEURT}  &  {ChrF++}  &  {BLEURT}  &  {CHR++} \\
\hline
Morpho  &  188  &  57.1  &  40.9  &  54.7  &  39.8  &  53.1  &  34.7  &  55.0  &  38.5 \\
 \text{ } + XPOS  &  190  &  57.7  &  41.1  &  55.6  &  39.9  &  54.1  &  35.3  &  55.8  &  38.8 \\
 \text{ } + BERT  &  190  &  59.4  &  42.5  &  57.1  &  40.4  &  56.1  &  36.3  &  57.5  &  39.7 \\
{\text{ } + BERT + XPOS}  &  192  &  \B{59.9}  &  \B{43.1}  &  \B{58.0}  &  \B{41.3}  &  \B{56.7}  &  \B{37.0}  &  \B{58.2}  &  \B{40.5} \\
\hline
\end{tabular}
}
}
\caption{Impact of attention augmentation: \B{Kinyarwanda $\rightarrow$ English}}
\label{table:attention_aug}
\vspace{-0.05in}
\end{table*}


\begin{table*}[!ht]
\centering
\resizebox{0.98 \textwidth}{!}{
{\renewcommand{\arraystretch}{1.3}
\begin{tabular}{| l | c c | c c | c c | c c |}
\hline
{~}  &  \multicolumn{2}{|c|}{FLORES-200} & \multicolumn{2}{|c|}{MAFAND-MT} & \multicolumn{2}{|c|}{TICO-19} & \multicolumn{2}{|c|}{\B{Average}} \\
\multicolumn{1}{|c|}{\B{Setup}}  &  {BLEURT}  &  {ChrF++}  &  {BLEURT}  &  {ChrF++}  &  {BLEURT}  &  {ChrF++}  &  {BLEURT}  &  {CHR++} \\
\hline

{Morpho + BERT + XPOS without Lexical Data}  &  58.5  &  41.6  &  56.3  &  39.9  &  55.5  &  36.0  &  56.8  &  39.2 \\
{Morpho + BERT + XPOS + Lexical Data}  &  \B{59.9}  &  \B{43.1}  &  \B{58.0}  &  \B{41.3}  &  \B{56.7}  &  \B{37.0}  &  \B{58.2}  &  \B{40.5} \\

\hline
\end{tabular}
}
}
\caption{Impact of lexical data (bilingual dictionaries): \B{Kinyarwanda $\rightarrow$ English}}
\label{table:lexical_aug}
\vspace{-0.05in}
\end{table*}


\begin{table*}[!ht]
\centering
\resizebox{0.98 \textwidth}{!}{
{\renewcommand{\arraystretch}{1.3}
\begin{tabular}{| l | c | c c | c c | c c | c c |}
\hline
{~}  &  \B{\#Params}  & \multicolumn{2}{|c|}{FLORES-200} & \multicolumn{2}{|c|}{MAFAND-MT} & \multicolumn{2}{|c|}{TICO-19} & \multicolumn{2}{|c|}{\B{Average}} \\
\multicolumn{1}{|c|}{\B{Setup}}  &  \B{ ($\times$ 1M)}  &  {BLEURT}  &  {ChrF++}  &  {BLEURT}  &  {ChrF++}  &  {BLEURT}  &  {ChrF++}  &  {BLEURT}  &  {CHR++} \\
\hline

{BPE Seq2Seq + XPOS}  &  187  &  50.1  &  35.5  &  48.5  &  34.2  &  47.4  &  30.7  &  48.7  &  33.5 \\
{Morpho + XPOS}  &  190  &  \B{57.7}  &  \B{41.1}  &  \B{55.6}  &  \B{39.9}  &  \B{54.1}  &  \B{35.3}  &  \B{55.8}  &  \B{38.8} \\

\hline
\end{tabular}
}
}
\caption{Impact of source side morphological modeling: \B{Kinyarwanda $\rightarrow$ English}}
\label{table:source_morpho}
\vspace{-0.05in}
\end{table*}


\begin{table*}[!ht]
\centering
\resizebox{0.98 \textwidth}{!}{
{\renewcommand{\arraystretch}{1.3}
\begin{tabular}{| l | c | c c | c c | c c | c c |}
\hline
~  &  ~ & \multicolumn{2}{|c|}{FLORES-200}  &  \multicolumn{2}{|c|}{MAFAND-MT}  &  \multicolumn{2}{|c|}{TICO-19}  &  \multicolumn{2}{|c|}{\B{Average}} \\
\multicolumn{1}{|c|}{\B{Setup}}  &  \B{\#Params}  & Dev  &  Test  &  Dev  &  Test  &  Dev  &  Test  &  Dev  &  Test \\
  ~ &  \B{ ($\times$ 1M)}  &  ChrF++  &  ChrF++  &  ChrF++  &  ChrF++  &  ChrF++  &  ChrF++  &  ChrF++  &  ChrF++ \\
\hline

{BPE Seq2Seq + XPOS} & 187 & 35.0 & 35.2 & 37.0 & 37.8 & 30.1 & 31.1 & 34.0 & 34.7 \\
{Morpho + XPOS (Loss summation)} & 196 & 36.9 & 37.2 & 39.2 & 40.9 & 32.0 & 33.0 & 36.0 & 37.0 \\
{Morpho + XPOS + GradVacc} & 196 & \B{37.6} & \B{38.2} & \B{41.0} & \B{42.4} & \B{32.8} & \B{33.5} & \B{37.1} & \B{38.0} \\

\hline
\end{tabular}
}
}
\caption{Impact of target side morphological modeling: \B{English $\rightarrow$ Kinyarwanda}}
\label{table:target_morpho}
\vspace{-0.05in}
\end{table*}


\begin{table*}[!ht]
\centering
\resizebox{0.98 \textwidth}{!}{
{\renewcommand{\arraystretch}{1.3}
\begin{tabular}{| l | r | c c | c c | c c |}
\hline
~  & ~  &  \multicolumn{2}{|c|}{FLORES-200} &  \multicolumn{2}{|c|}{MAFAND-MT}  &  \multicolumn{2}{|c|}{TICO-19} \\
\multicolumn{1}{|c|}{\B{Setup}} & \B{\#Params} &  Dev  &  Test  &  Dev  &  Test  &  Dev  &  Test \\
  ~ & \B{x 1M} &  chrF2  &  chrF2  &  chrF2  &  chrF2  &  chrF2  &  chrF2 \\
\hline
\U{Morpho + XPOS + BERT + Backtransl. (\B{Ours})}     &  403       &  \B{53.2}  &  \B{53.1}  &  \B{58.2}  &  \B{61.9}  &  48.7*  &  50.2* \\
{Helsinki-opus-mt~\cite{tiedemann2020opus}}      &  76             &  35.5  &  36.7  &  34.3  &  37.3  &  27.5  &  27.2 \\
{NLLB-200 600M (distilled)~\cite{costa2022no}}  &  600             &  45.8  &  45.5  &  50.4  &  52.7  &  44.8  &  46.3 \\
mBART~\cite{liu2020multilingual} fine-tuned on our dataset &  610  &  48.7  &  48.5  &  52.4  &  54.1  &  43.7  &  45.2 \\
{NLLB-200 3.3B~\cite{costa2022no}}             &  3,300            &  50.6  &  50.9  &  \B{57.3}  &  58.6  &  \B{\U{50.0}}  &  \B{\U{52.4}} \\
\hline
{Google Translate}                              &  N/A             &  \U{59.1}  &  \U{60.0}  &  \U{76.6}  &  \U{87.5}  &  46.5  &  49.6 \\
\hline
\end{tabular}
}
}
\caption{\B{English $\rightarrow$ Kinyarwanda}: Comparison of our large model performance after back-translation in relation to open-source models and Google Translate. chrF2 scores are computed using SacreBLEU~\citep{post2018call} with 10000 bootstraps for significance testing. Highest scores among open source models (p-value < 0.002) are shown in bold. Overall best scores are underlined. *On TICO-19, our model outperforms Google Translate (p-value < 0.002).}
\label{table:back_trans_eng_kin}
\vspace{-0.15in}
\end{table*}

\begin{table*}[!ht]
\centering
\resizebox{0.98 \textwidth}{!}{
{\renewcommand{\arraystretch}{1.3}
\begin{tabular}{| l | r | c c | c c | c c |}
\hline
~  & ~  &  \multicolumn{2}{|c|}{FLORES-200} &  \multicolumn{2}{|c|}{MAFAND-MT}  &  \multicolumn{2}{|c|}{TICO-19} \\
\multicolumn{1}{|c|}{\B{Setup}} & \B{\#Params} &  Dev  &  Test  &  Dev  &  Test  &  Dev  &  Test \\
  ~ & \B{x 1M} &  chrF2  &  chrF2  &  chrF2  &  chrF2  &  chrF2  &  chrF2 \\
\hline
\U{Morpho + XPOS + BERT + Backtransl. (\B{Ours})}     &  396       &  54.6  &  54.8  &  \B{54.7}  &  \B{59.2}  &  49.4  &  50.1 \\
{Helsinki-opus-mt~\cite{tiedemann2020opus}}      &  76             &  35.4  &  35.1  &  33.7  &  35.1  &  29.6  &  29.7 \\
{NLLB-200 600M (distilled)~\cite{costa2022no}}  &  600             &  53.1  &  52.3  &  51.1  &  54.9  &  47.7  &  48.6 \\
mBART~\cite{liu2020multilingual} fine-tuned on our dataset &  610  &  43.7  &  43.1  &  44.5  &  46.0  &  38.8  &  38.8 \\
{NLLB-200 3.3B~\cite{costa2022no}}             &  3,300            &  \B{56.8}  &  \B{56.0}  &  \B{55.4}  &  \B{59.6}  &  \B{\U{53.4}}  &  \B{\U{54.1}} \\
\hline
{Google Translate}                              &  N/A             &  \U{60.0}  &  \U{59.1}  &  \U{57.3}  &  \U{64.0}  &  51.8  &  52.4 \\
\hline
\end{tabular}
}
}
\caption{\B{Kinyarwanda $\rightarrow$ English}: Comparison of our large model performance after back-translation in relation to open-source models and Google Translate. chrF2 scores are computed using SacreBLEU~\citep{post2018call} with 10000 bootstraps for significance testing. Highest scores among open source models (p-value < 0.002) are shown in bold. Overall best scores are underlined.}
\label{table:back_trans_kin_eng}
\vspace{-0.15in}
\end{table*}

\subsection{Evaluation}

We evaluate our models on three different benchmarks that include Kinyarwanda, namely FLORES-200~\cite{costa2022no}, MAFAND-MT~\cite{adelani2022few} and TICO-19~\cite{anastasopoulos2020tico}. This allow us to have a picture on how the models perform on different domains, respectively Wikipedia, News and Covid-19. Our main evaluation metric is ChrF++~\cite{popovic2017chrf++} which includes both character-level and word-level n-gram evaluation, does not rely to any sub-word tokenization and has been shown to correlate better with human judgements than the more traditional BLUE score. We use TorchMetrics~\cite{detlefsen2022torchmetrics} package's default implementation of ChrF++. For Kinyarwanda$\rightarrow$ English translation, we also evaluate with BLEURT scores~\cite{sellam2020bleurt}, an embedding-based metric with higher correlation with human judgement. We use a pre-trained PyTorch implementation of BLEURT\footnote{\url{https://github.com/lucadiliello/bleurt-pytorch}}. We did not use BLEURT scores for English$\rightarrow$Kinyarwanda because there was no available pre-trained BLEURT model for Kinyarwanda and the pre-training cost is very high.

The baseline BPE-based models in~\tblref{source_morpho} and~\tblref{target_morpho} use a SentencePiece~\cite{kudo2018sentencepiece} tokenizer, with 32K-token vocabularies for either source or target. The SentencePiece tokenizers are trained/optimized on 16M sentences of text for each language. Source and target vocabularies are not shared. The NMT models in this case use the same Transformer backbone as the morphological models, but without morphological modeling or BERT attention augmentation.


\subsection{Results}

Results in \tblref{name_copy} through \tblref{target_morpho} show our ablation study results, evaluating the various contributions. In \tblref{name_copy}, we show the improvement across all three benchmarks from adding proper names data to induce token-copying ability. In \tblref{attention_aug}, we evaluate the impact of our attention augmentation scheme. The results show substantial improvement by adding BERT and XPOS attention augmentations. \tblref{lexical_aug} confirms the effectiveness of adding bilingual dictionary data to the training set. In \tblref{source_morpho} and \tblref{target_morpho}, we find a large performance gap between standard transformer with BPE-based tokenization (BPE Seq2Seq) and our morphology-based models (Morpho), which confirms the effectiveness of our morphological modeling. Finally, in \tblref{back_trans_eng_kin} and \tblref{back_trans_kin_eng}, we use back-translation and train 400M-parameter models that perform better than strong baselines including NLLB-200 (3.3B parameters for English$\rightarrow$ Kinyarwanda, 600M parameters for both directions) and fine-tuned mBART (610M parameters). For English$\rightarrow$Kinyarwanda, we achieve performance exceeding that of Google Translate on the out-of-domain TICO-19 benchmark.















\section{Related Work}

Morphological modeling in NMT is an actively researched subject often leading to improvements in translation. However most of this research has focused on European languages. \citet{ataman2018compositional} shows that an RNN-based word compositional model improves NMT on several languages. \citet{weller2020modeling} evaluates both source-side and target-side morphology modeling between English and German using a lemma+tag representation. \citet{passban2018improving} proposes using multi-task learning of target-side morphology with a weighted average loss function. However, \citet{machavcek2018morphological} does not find improvement when using unsupervised morphological analysers. Our studies differs in that it uses a different morphological representation, that is the two-tier architecture, and we also evaluate on a relatively lower resourced language.

The idea of model augmentation with pre-trained language models (PLM) have been previously explored by \citet{sun2021multilingual}, and \citet{zhu2020incorporating} who use a drop-net scheme to integrate BERT embeddings. Also, there have been attempts to model word order relationships between source and target languages~\cite{li2017modeling, murthy2019addressing}. Our model architecture provides a more generic approach through the attention augmentation.

\section{Conclusion}
This work combines three techniques of morphological modeling, attention augmentation and data augmentation to improve machine translation performance for a low-resource morphologically-rich language. Our ablation results indicate improvement from each individual contribution. Baseline improvements from morphological modeling are more pronounced at the target side than at the source side. This work expands the landscape of modeling complex morphology in NMT and provides a potential framework-solution for machine translation of low-resource morphologically rich languages.

\section{Limitations}

Our morphological modeling proposal requires an effective morphological analyzer and was only evaluated on one morphologically-rich language, that is Kinyarwanda. Morphological analyzers are not available for all languages and this will limit the applicability of our technique. The proposed data augmentation technique for enabling proper name copying ability works in most cases, but we also observed some few cases where inexact copies are produced. Similarly, even with lexical data added to our training, we still observe some cases of hallucinated output words, mostly when the model encounters unseen words. Finally, the proposed morphological decoding algorithm is slower than standard beam search because of the filtering steps and morphological synthesis performed before producing a candidate output token.

Given the above limitations, our model does not grant complete reliability and the produced translations still require post-editing to be used in high-stake applications. However, there are no major risks for using the model in normal use cases as a translation aid tool.

\bibliography{anthology,custom}

\appendix

\section{Model, training and inference hyper-parameters}
\label{sec:hyper_params}

Our model hyper-parameters along with training and inference hyper-parameters are provided in~\tblref{hyper_parameters}.

\begin{table}[!ht]
\centering
\resizebox{0.92 \columnwidth}{!}{
{\renewcommand{\arraystretch}{1.3}
\begin{tabular}{| l | r |}
\hline
\B{Hyper-parameter} & \B{Value} \\
\hline
\B{Model} & \\
Transformer hidden dimensions & 768 \\
Transformer feed-forward dimension & 3072 \\
Transformer attention heads & 12 \\
Transformer encoder layers with BERT & 5 \\
Transformer encoder layers without BERT & 8 \\
Transformer decoder layers with GPT & 7 \\
Transformer decoder layers without GPT & 8 \\
\textit{Morpho-Encoder} hidden dimension & 128 \\
\textit{Morpho-Encoder} feed-forward dimension & 512 \\
\textit{Morpho-Encoder} attention heads & 4 \\
\textit{Morpho-Encoder} layers & 3 \\
Dropout & 0.1 \\
Maximum sequence length & 512 \\
\hline
\B{Training} & \\
Batch size & 32K tokens \\
Peak learning rate & 0.0005 \\
Learning rate schedule & inverse sqrt \\
Warm-up steps & 8000 \\
Optimizer & Adam \\
Adam's $\beta_1$, $\beta_2$ & 0.9, 0.98 \\
Maximum training epochs & 40 \\
\hline
\B{Morphological inference} & \\
Beam width & 4 \\
Top scores $M$ & 8 \\
Cut-off gap $\delta$ & 0.3 \\
Minimum affix probability $\gamma$ & 0.3 \\
Global correlation weight $\alpha$ & 0.08 \\
Surface form score stop gap $log(\beta)$ & 2.0 \\
\hline
\end{tabular}
}
}
\caption{Hyper-parameter settings}
\label{table:hyper_parameters}
\end{table}

\newpage
\section{Dataset summary}
\label{sec:data_summary}

\begin{table}[!ht]
\centering
\resizebox{0.9 \columnwidth}{!}{
{\renewcommand{\arraystretch}{1.3}
\begin{tabular}{| l | l |}
\hline
\B{Subset} & \B{Size} \\
\hline
\B{Parallel sentences} & ~\\
Jw.org website & 562,417 sentences \\
Rwanda's official gazette & 113,127 sentences \\
\hline
\B{Lexical data} & ~ \\
Iriza dictionary 2006 & 108,870 words \\
Kinyarwanda.net & 10,653 words \\
Manually translated & 8,000 words \\
\hline
\B{Augmented data} & ~ \\
Spelled numbers & 200,000 phrases \\
Copy data (e.g. Proper names) & 157,668 words \\
Code-switching foreign terms & 10,276 terms \\
\hline
\end{tabular}
}
}
\caption{Summary of our experimental parallel dataset}
\label{table:dataset_size}
 \vspace{-0.1in}
\end{table}


\newpage
\onecolumn
\section{Morphological decoding algorithm}
\label{sec:morpho_decoding}

\begin{algorithm*}[!h]
    \SetKwFunction{getInflectionGroup}{getInflectionGroup}
    \SetKwFunction{min}{min}
    \SetKwFunction{max}{max}
    \SetKwFunction{array}{array}
    \SetKwFunction{log}{log}
    \SetKwFunction{ln}{ln}
    \SetKwFunction{of}{of}
    \SetKwFunction{size}{size}
    \SetKwFunction{softmax}{softmax}
    \SetKwFunction{argSort}{argSort}
    \SetKwFunction{inflectionGroupProbabilities}{inflectionGroupProbabilities}
    \SetKwFunction{filterAndCutOff}{filterAndCutOff}
    \SetKwFunction{computeScore}{computeScore}
    \SetKwFunction{generateInflections}{generateInflections}
    

    
    \SetKwProg{myalg}{Subroutine:}{}{}

    \myalg{$\inflectionGroupProbabilities(T_s, T_p, T_a, P_s, P_p, P_a, M, G)${}}{
    
    {$W \gets M\times G$ \array}

    {$W[i,g] \gets -100 $,~$ \forall i=1,2,..M, \forall g=1,2,..G$}

    \For{$i \leftarrow 1$ \KwTo $M$}{
        
        {$g_s \gets \getInflectionGroup(T_s[i]$)}
        
        {$W[i,g_s] \gets \max(W[i,g_s], \log(P_s[T_s[i]]$))}

        {$g_p \gets \getInflectionGroup(T_p[i]$)}
        
        {$W[i,g_p] \gets \max(W[i,g_p], \log(P_p[T_p[i]]$))}

        {$g_a \gets \getInflectionGroup(T_a[i]$)}
        
        {$W[i,g_a] \gets \max(W[i,g_a], \log(P_a[T_a[i]]$))}
    }

    {$V \gets G$ \array}

    {$V[g] \gets \sum_{i=1}^{M}{W[i,g]}, \forall g=1,2,..G$}

    {$P_g \gets \softmax(V)$}
    
    \KwRet{$P_g$}
    }

    \myalg{$\filterAndCutOff(T,P,g,\delta,\gamma)${}}{
    
    {$T^{'} \gets [~]$}

    {$pp \gets 0$}

    \ForEach{$i \in T$}{
        \uIf{($\getInflectionGroup(i) = g$) and ($P[i] \geq \gamma$)}{
            {$T^{'}.append(i)$}
            
            \uIf{$(pp - P[i]) > \delta$} {
                {break}
            }
            {$pp \gets P[i]$}
        }
    }
    \KwRet{$T^{'}$}
    }

    \myalg{$\computeScore(T_s, T_p, T_a, P_s, P_p, P_a,\rho[.,.],\alpha)${}}{
    
    {$C \gets |T_s|\times|T_p|\times|T_a|$ \array}
    
    {$tot \gets 0$}
    
    \ForEach{$s \in Ts$}{
        \ForEach{$p \in Tp$}{
            \ForEach{$a \in Ta$}{
                {$c \gets \exp(\alpha \log(\rho[s,a]) + \log P_s[s] + \log P_p[p] + \log P_a[a])$}

                {$C[s,p,a] \gets c$}

                {$tot \gets tot+c$}
            }
        }
    }
    {$C[s,p,a] \gets C[s,p,a]/tot $;~$ \forall s\in T_s, \forall p\in T_p, \forall a\in T_a$} \tcp{Normalize}
    
    \KwRet{$C$}
    }
    \caption{Inflection generation helper subroutines}
\end{algorithm*}

\begin{algorithm*}
    \SetKwFunction{getInflectionGroup}{getInflectionGroup}
    \SetKwFunction{min}{min}
    \SetKwFunction{max}{max}
    \SetKwFunction{array}{array}
    \SetKwFunction{log}{log}
    \SetKwFunction{ln}{ln}
    \SetKwFunction{of}{of}
    \SetKwFunction{size}{size}
    \SetKwFunction{softmax}{softmax}
    \SetKwFunction{null}{null}
    \SetKwFunction{argSort}{argSort}
    \SetKwFunction{affixMerge}{affixMerge}
    \SetKwFunction{morphoSynthesis}{morphoSynthesis}
    \SetKwFunction{inflectionGroupProbabilities}{inflectionGroupProbabilities}
    \SetKwFunction{filterAndCutOff}{filterAndCutOff}
    \SetKwFunction{computeScore}{computeScore}
    \SetKwFunction{generateInflections}{generateInflections}
    
    \SetKwInOut{KwIn}{Input}
    \SetKwInOut{KwOut}{Output}

    \KwIn{Probability distributions returned by the neural network (softmax heads) for stems, POS tags, affix sets ($P_s, P_p, P_a$); probability values returned for each affix (multi-label heads) $P_f$; $M$: number of top items (stems, POS tags, affix sets) to consider; $N$: number of top affixes to consider; number of inflection groups $G$; $\rho[.,.]$: corpus-computed correlation of stems and affix sets; $\alpha$: stem-affix set correlation weight; $\beta$: maximum probability gap for inflection groups; $\gamma$: minimum affix probability; $\delta$: maximum probability gap for stems, POS tags and affix sets.}
    \KwOut{Candidate inflected forms and their scores.}
    
    \SetKwProg{myalg}{Subroutine}{}{}

    \myalg{$\generateInflections(P_s, P_p, P_a, P_f,\rho[.,.],\alpha,\beta,\gamma,\delta)${}}{
    
    \tcc{All $\argSort(.)$ calls are in decreasing order}
    {$T_s \gets \argSort(P_s)[:M]$} \tcp{Up to $M$ items}

    {$T_p \gets \argSort(P_p)[:M]$}
    
    {$T_a \gets \argSort(P_a)[:M]$}
    
    {$T_f \gets \argSort(P_f)[:N]$}
    
    {$P_g \gets \inflectionGroupProbabilities(T_s, T_p, T_a, P_s, P_p, P_a, M, G)$}

    {$G_s \gets \argSort(P_g)$} 

    {$pp \gets 0$}

    {$R \gets [~]$} \tcp{List of inflections to return}
    
    \ForEach{$g \in G_s$} {
        {$T_s^{'} \gets \filterAndCutOff(T_s,P_s,g,\delta,0)$}
        
        {$T_p^{'} \gets \filterAndCutOff(T_p,P_p,g,\delta,0)$}
        
        {$T_a^{'} \gets \filterAndCutOff(T_a,P_a,g,\delta,0)$}
        
        {$T_f^{'} \gets \filterAndCutOff(T_f,P_f,g,\delta,\gamma)$}
        
        {$C_g \gets \computeScore(T_s^{'}, T_p^{'}, T_a^{'}, P_s, P_p, P_a,\rho[.,.],\alpha)$}
        
        {$L_g \gets \argSort(C_g)$}
        
        \ForEach{$(s,p,a) \in L_s$} {
             \tcp{Formulate affixes by merging affix set's }
             \tcp{own affixes and extra predicted affixes}
            {$f \gets \affixMerge(a,T_f^{'})$}

            \tcp{Call morphological synthesizer}
            ${surface \gets \morphoSynthesis(s, f)}$

            \uIf{$surface \neq \null$} {
                {$R.append((surface,s,p,a,f,C_g[s,p,a]))$}
            }
            
        }
        \uIf{$(pp - P_g[g]) > \beta$} {
            {break}
        }
        {$pp \gets P_g[g]$}
    }
    \KwRet{$R$}
    
    \tcp{The returned inflections will be added to the beam }
    \tcp{for beam search-based decoding.}
    }
    \caption{Inflection generation}
\end{algorithm*}

\end{document}